\documentclass[sigconf]{acmart}
\AtBeginDocument{%
  }

\setcopyright{acmlicensed}
\copyrightyear{2018}
\acmYear{2018}
\acmDOI{XXXXXXX.XXXXXXX}
\acmConference[Conference acronym 'XX]{Make sure to enter the correct
  conference title from your rights confirmation email}{June 03--05,
  2018}{Woodstock, NY}
\acmISBN{978-1-4503-XXXX-X/2018/06}

\usepackage{booktabs}
\usepackage[ruled,linesnumbered]{algorithm2e}
\usepackage{microtype}
\usepackage{graphicx}
\usepackage{subcaption}
\usepackage{balance}
\usepackage{hyperref}
\usepackage{amsmath}
\usepackage{amssymb}
\usepackage{mathtools}
\usepackage{amsthm}
\usepackage{enumitem}
\usepackage{xcolor}
\usepackage{multirow}
\usepackage{float}

\usepackage{amsthm}

\newtheorem{assumption}{Assumption}

\newcommand{\R}{\mathbb{R}}

\setcopyright{none}
\acmDOI{}
\acmISBN{}
\acmConference[Arxiv]{Preprint}
{April 2026}{Preprint}
\acmYear{}
\copyrightyear{}

\begin{document}

\title{Federated Cross-Modal Retrieval with Missing Modalities via \\ Semantic Routing and Adapter Personalization}

\author{Hefeng Zhou}
\authornote{Both authors contributed equally to this research.}
\email{zadeji1@sjtu.edu.cn}
\author{Xuan Liu}
\authornotemark[1]

\author{Sicheng Chen}
\affiliation{%
  \institution{Shanghai Jiao Tong University}
  \city{}
  \state{}
  \country{}
}
\affiliation{%
  \institution{Shanghai Jiao Tong University}
  \city{}
  \state{}
  \country{}
}
\affiliation{%
  \institution{Hohai University}
  \city{}
  \state{}
  \country{}
}

\author{Wutong Zhang}
\author{Yan Wu}
\author{Jiong Lou}
\author{Chentao Wu}
\author{Guangtao Xue}

\affiliation{%
  \institution{Shanghai Jiao Tong University}
  \city{}
  \country{}
}

\author{Wei Zhao}
\author{Jie Li}
\authornote{Corresponding author}
\affiliation{%
  \institution{Shenzhen Univ of Advanced Technology}
  \city{}
  \country{}
}
\email{lijiecs@sjtu.edu.cn}
\affiliation{%
  \institution{Shanghai Jiao Tong University}
  \city{}
  \country{}
}

\renewcommand{\shortauthors}{Zhou et al.}

\begin{abstract}
Federated cross-modal retrieval faces severe challenges from heterogeneous client data, particularly non-IID semantic distributions and missing modalities. Under such heterogeneity, a single global model is often insufficient to capture both shared cross-modal knowledge and client-specific characteristics. We propose RCSR, a personalization-friendly federated framework that integrates prototype anchoring, retrieval-centric semantic routing, and optional client-specific adapters. Built on a frozen CLIP backbone, RCSR leverages lightweight shared adapters for global knowledge transfer while supporting efficient local personalization. Prototype anchoring helps unimodal clients align with global cross-modal semantics, and a server-side semantic router adaptively assigns aggregation weights based on retrieval consistency to mitigate alignment drift during heterogeneous updates. Extensive experiments on MS-COCO, Flickr30K, and other benchmarks show that RCSR consistently improves global retrieval accuracy and training stability, while further enhancing client-level retrieval performance, especially for clients with incomplete modalities. Code is available at \url{https://github.com/RezinChow/RCSR-Retrieval-Centric-Semantic-Routing}.
\end{abstract}



\begin{teaserfigure}
    \centering
  \includegraphics[width=0.95\textwidth]{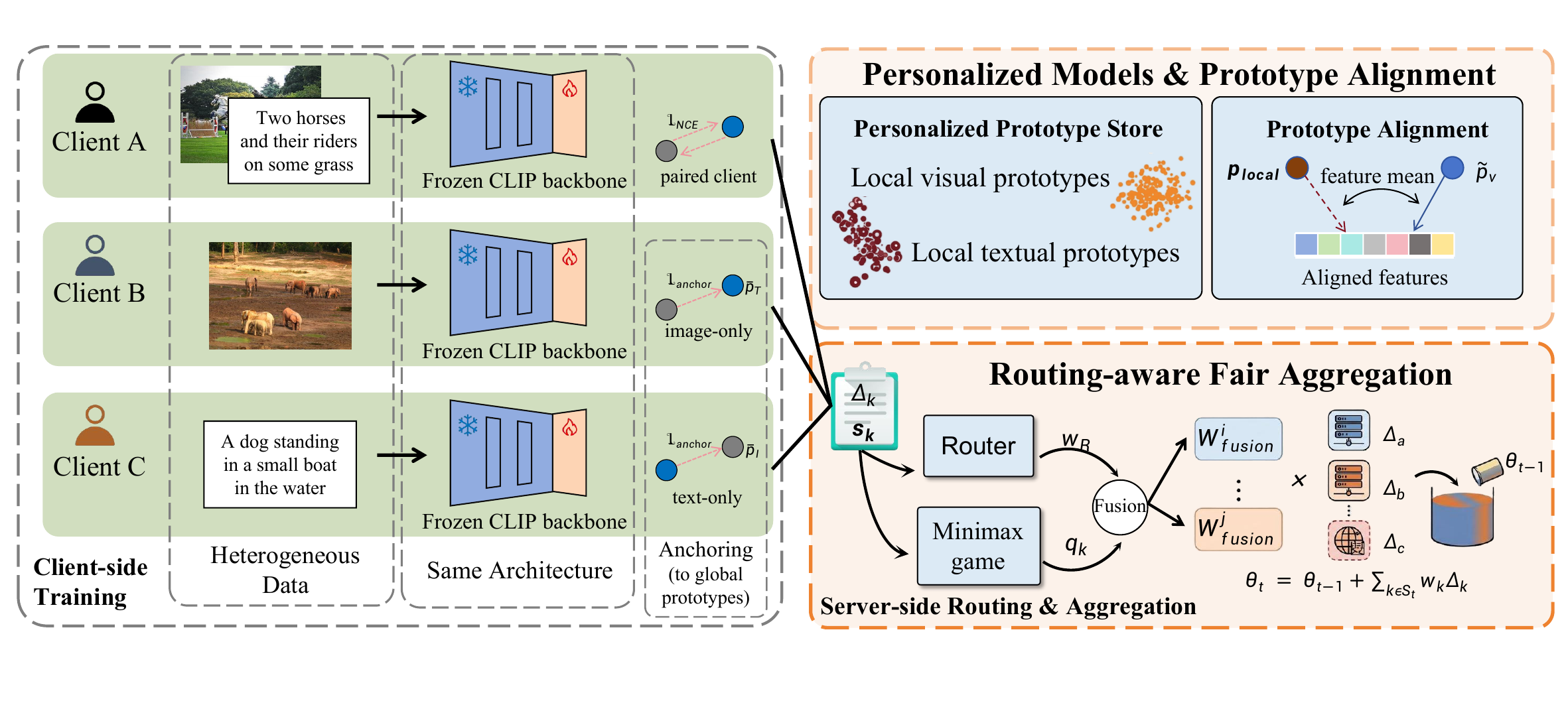}
  \caption{Our method: a cross-modal solution for scenarios with missing modalities.}
  \Description{Our method: a cross-modal solution for scenarios with missing modalities.}
  \label{fig:teaser}
\end{teaserfigure}

\maketitle

\section{Introduction}

The widespread availability of images and text has made multimodal retrieval a key component in modern information systems, where a query in one modality is used to retrieve semantically relevant items in another modality \cite{Baltruvsaitis2018MultimodalSurvey}. High-performing retrieval models typically depend on large-scale and diverse training data, yet such data are often dispersed across organizations and personal devices. Privacy regulations and data governance further restrict centralized collection, creating persistent data silos that limit model quality and coverage. Federated learning offers a practical alternative by enabling collaborative training without transferring raw data, thereby improving utility while respecting privacy constraints \cite{Yang2019FMLConcept,Li2021FLSurvey}.

Despite this promise, directly deploying federated learning for cross-modal retrieval introduces challenges that are less prominent in unimodal classification. Real clients frequently have incomplete modalities, where some participants only observe images and others only observe text, which breaks standard contrastive training that relies on paired data \cite{wu2024deep}. Strong statistical heterogeneity across clients can distort the shared embedding space, making it difficult to maintain consistent alignment across modalities at the global level \cite{Xiong2022UnifiedMFL}. Even when raw data stay local, the exchanged training signals can still raise privacy concerns, motivating designs that are communication efficient and privacy conscious, particularly in sensitive applications that already demand privacy-preserving multimodal modeling \cite{gao2026privacy}.

To address heterogeneity and reduce communication burdens, recent federated systems increasingly explore knowledge distillation and representation level communication as alternatives to full parameter exchange \cite{Lin2020FedDF}. Distillation oriented federated learning has also been studied from a broader practical perspective, highlighting when and how compact knowledge transfer can improve scalability \cite{Mora2022DistillationFL,He2020FedGKT}. However, for cross-modal retrieval, a central difficulty remains underexplored: how to preserve retrieval relevant geometry when clients participate with missing modalities and non-IID data, while keeping communication lightweight and training stable. This motivates a retrieval centric federated perspective that treats alignment consistency as a first class objective and supports fragmented modality participation in both training and evaluation.

We propose Retrieval-Centric Semantic Routing (RCSR) to enable federated cross-modal retrieval under non-IID data and missing modalities. RCSR uses prototype anchoring to train single-modality clients and a retrieval-centric router with robust-fair minimax aggregation to stabilize alignment and improve worst-case performance. Experiments on Flickr30K, MS-COCO and other datasets show consistent gains in accuracy and stability under severe incompleteness.

\section{Related Work}
\label{sec:related-work}

\subsection{Cross-Modal Retrieval}

Cross-modal retrieval learns a shared embedding space for bidirectional image--text matching.
Early methods adopt dual-encoder architectures with ranking losses \cite{faghri2017vse++}, while cross-attention models capture fine-grained region--word correspondences \cite{lee2018stacked,wang2020pfan++}.
Large-scale vision--language pre-training~\cite{du2022survey} has since shifted the paradigm toward contrastive dual encoders, with CLIP \cite{radford2021learning} and ALIGN \cite{jia2021scaling} demonstrating strong zero-shot transferability.
We build on CLIP as the frozen backbone and fine-tune lightweight adapters for federated retrieval, which has received limited attention.

\subsection{Federated Learning with Heterogeneity}

Standard FedAvg \cite{mcmahan2017communication} suffers under non-IID data and partial participation.
FedProx \cite{li2020federated} adds a proximal regularizer, SCAFFOLD \cite{karimireddy2020scaffold} corrects client drift via control variates, and FedNova \cite{wang2020tackling} normalizes updates for variable local computation.
Representation-level approaches exchange compact signals: FedProto \cite{tan2022fedproto} communicates class prototypes, MOON \cite{li2021model} applies model-level contrastive learning, and FedPer \cite{arivazhagan2019federated} splits global and personalized layers.
For fairness, AFL \cite{mohri2019agnostic} optimizes worst-case performance via minimax reweighting, and q-FFL \cite{li2019fair} redistributes effort toward disadvantaged clients.
Personalization methods such as pFedMe \cite{t2020personalized} and pFedMMA \cite{ghiasvand2025pfedmma} learn client-adapted parameters on top of a shared model.

\subsection{Federated Multimodal Learning}

Federated multimodal learning faces additional challenges from missing modalities and cross-modal alignment drift.
FedCMR \cite{zong2021fedcmr} aggregates a shared cross-modal subspace; CreamFL \cite{yu2023multimodal} uses representation-level ensemble distillation with contrastive regularization.
MFCPL \cite{le2025cross} constructs cross-modal prototypes for knowledge transfer under severe missing modalities, and FedMEKT \cite{le2025fedmekt} transfers embedding-level knowledge via distillation.
However, existing methods either rely on generic aggregation or public-data-driven distillation, leaving underexplored how to explicitly route aggregation toward retrieval-centric geometric consistency when clients simultaneously face non-IID data, missing modalities, and limited communication budgets.
This gap motivates our retrieval-centric semantic routing with minimax fairness.

\section{Problem Formulation}
\label{sec:problem}

\textbf{Setup.}
Consider $N$ clients, each holding a local dataset $\mathcal{D}_k$ of image--text pairs partitioned via $\mathrm{Dir}(\alpha)$~\cite{hsu2019measuring} over class labels. A subset $\mathcal{S}_t \subseteq [N]$ participates per round with rate $C$. The goal is to learn a dual-encoder model $\theta$ (image encoder $f_I$ and text encoder $f_T$) that maps both modalities into a shared embedding space $\R^d$ for cross-modal retrieval, evaluated by Recall@$K$.

\textbf{Missing Modality.}
Under a missing modality rate $\rho$, a fraction $\rho$ of clients possess only a single modality (image-only or text-only), denoted by mask $m_k$. These clients cannot compute standard cross-modal contrastive losses, breaking conventional federated training pipelines.

\textbf{Objective.}
We jointly optimize retrieval quality and worst-case fairness:
\begin{equation}\label{eq:objective}
\min_{\theta}\; \max_{q \in \Delta^{N-1}} \;\sum_{k=1}^{N} q_k \, \mathcal{L}_k(\theta) \;+\; \tau \, H(q),
\end{equation}
where $\mathcal{L}_k(\theta)$ is client $k$'s local loss, $q$ is an adversarial client weighting that exposes worst-case performance, and $H(q)$ is an entropy regularizer. To solve this under missing modalities, Section~\ref{sec:method} introduces: \emph{(i)} prototype anchoring for single-modality clients, \emph{(ii)} a retrieval-centric router $R_\phi$ producing adaptive weights $w^R$, and \emph{(iii)} fused aggregation $w_k \propto w_k^R \cdot q_k$.

\section{Methodology}
\label{sec:method}

\subsection{Overview}

RCSR decomposes federated cross-modal retrieval into three mechanisms: modality-aware local training with prototype anchoring for single-modality clients (\S\ref{sec:client}), a retrieval-centric semantic router that computes adaptive aggregation weights (\S\ref{sec:sec:router}), and robust-fair aggregation via a minimax game (\S\ref{sec:fairness}). Algorithm~\ref{alg:rcsr} details the protocol.

\begin{figure*}
    \centering
    \includegraphics[width=1\linewidth]{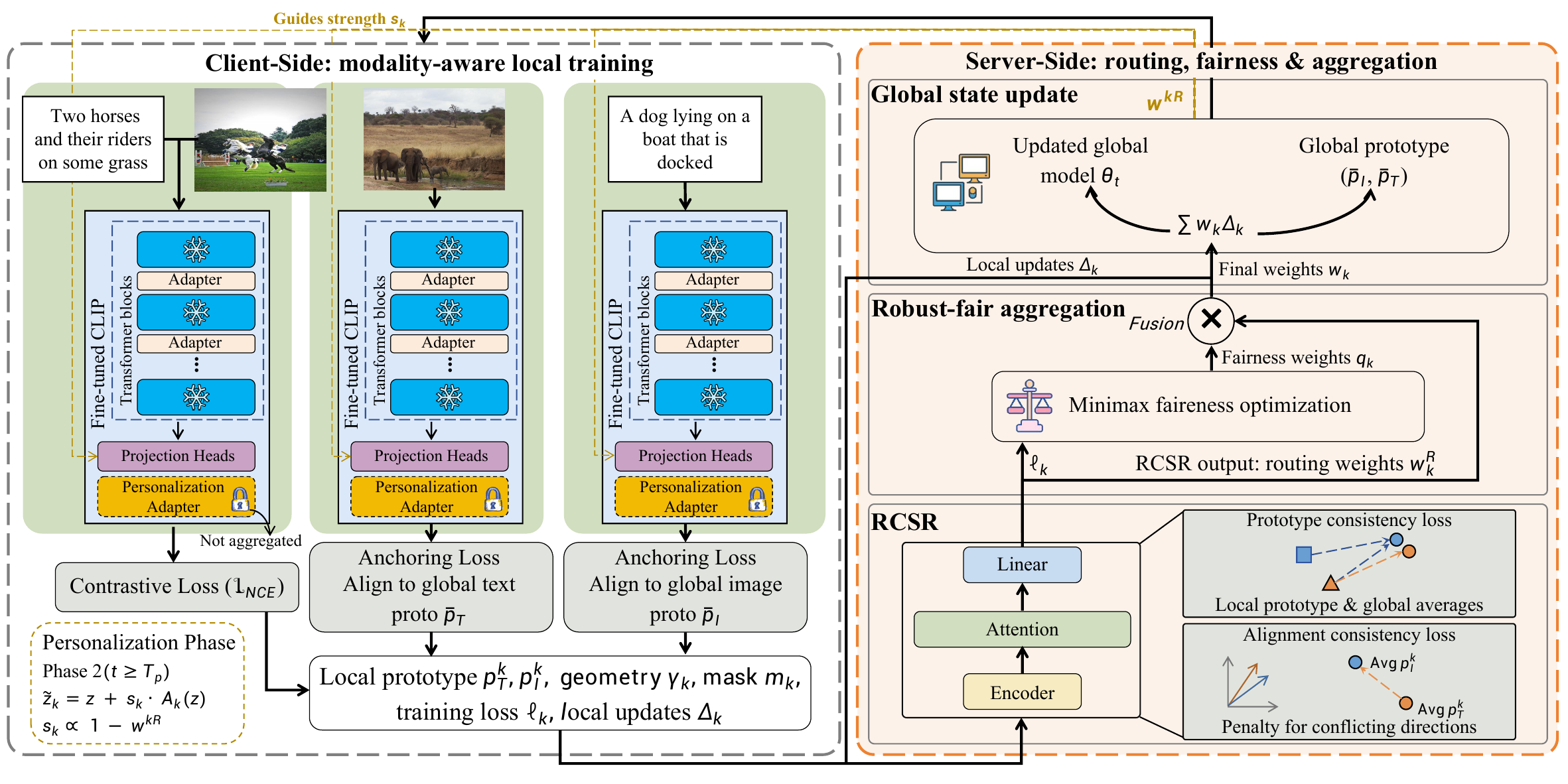}
    \caption{RCSR framework: client-side modality-aware training with prototype anchoring, server-side retrieval-centric routing with robust-fair aggregation, and optional personalization.}
    \label{fig:framework}
\end{figure*}

\begin{algorithm}[t]
\caption{RCSR Federated Training}
\label{alg:rcsr}
\small
\KwIn{$N$ clients $\{(\mathcal{D}_k, m_k)\}_{k=1}^N$, rounds $T$, warm-up $T_w$, fraction $C$}
\KwOut{Global model $\theta_T$}
\textbf{Init:} model $\theta_0$, router $R_\phi$, $q \leftarrow \mathbf{1}/N$, prototypes $\bar{p}_I, \bar{p}_T$\;
\For{$t = 1, \ldots, T$}{
    Sample $\mathcal{S}_t \subseteq [N]$, $|\mathcal{S}_t| = \lfloor CN \rfloor$\;
    Broadcast $\theta_{t-1}, \bar{p}_I, \bar{p}_T$ to selected clients\;
    \tcp{Phase 1: Local Training}
    \ForEach{$k \in \mathcal{S}_t$}{
        $\theta_k, \ell_k \leftarrow$ \textsc{LocalTrain}$(\theta_{t-1}, \mathcal{D}_k, m_k, \bar{p}_I, \bar{p}_T)$\;
        Compute $s_k$: available-modality prototypes + global fallback, geometry, mask, loss\;
    }
    \tcp{Phase 2: Routing (active after warm-up)}
    \eIf{$t > T_w$}{
        $w^R \leftarrow \mathrm{softmax}(R_\phi(\{s_k\}))$ \tcp*{Eq.~(\ref{eq:router_weights})}
        $\phi \leftarrow \phi - \alpha_R \nabla_\phi \mathcal{L}_R$ with $\mathrm{sg}[\bar{p}_v]$ \tcp*{stop-grad on EMA}
    }{
        $w^R \leftarrow |\mathcal{D}_k| / \sum_j |\mathcal{D}_j|$ \tcp*{FedAvg during warm-up}
    }
    \tcp{Phase 3: Fairness}
    $\tilde{\ell}_k \leftarrow \ell_k / \bar{\ell}_{m_k}$ per client type; $q_k \leftarrow q_k \cdot e^{\eta_q \tilde{\ell}_k}$, renorm $q$\;
    $w_k \leftarrow w_k^R q_k / \sum_{j} w_j^R q_j$ \tcp*{Eq.~(\ref{eq:fusion})}
    $\theta_t \leftarrow \theta_{t-1} + \sum_{k \in \mathcal{S}_t} w_k \Delta_k$\;
    $\bar{p}_v \leftarrow \mu \bar{p}_v + (1\!-\!\mu)\sum_k w_k p_v^k$, $v \in \{I,T\}$\;
}
\Return $\theta_T$\;
\end{algorithm}

\subsection{Client-Side Local Training}
\label{sec:client}

Each selected client $k$ receives the global model $\theta$ and global prototypes $(\bar{p}_I, \bar{p}_T)$, then trains with a modality-aware composite loss.

\textbf{Contrastive Learning.} Paired clients optimize symmetric InfoNCE:
\begin{equation}\label{eq:infonce}
\mathcal{L}_{\mathrm{NCE}} = -\frac{1}{2B}\sum_{i=1}^{B}\!\left[\log\frac{e^{z_I^i \cdot z_T^i/\tau_{\mathrm{NCE}}}}{\sum_j e^{z_I^i \cdot z_T^j/\tau_{\mathrm{NCE}}}} + \log\frac{e^{z_T^i \cdot z_I^i/\tau_{\mathrm{NCE}}}}{\sum_j e^{z_T^i \cdot z_I^j/\tau_{\mathrm{NCE}}}}\right],
\end{equation}
where $z_I, z_T \in \R^d$ are L2-normalized embeddings.

\textbf{Prototype Anchoring.} Single-modality clients cannot compute contrastive loss. We anchor their representations to the global prototype of the missing modality:
\begin{equation}\label{eq:anchor}
\mathcal{L}_{\mathrm{anchor}} = 1 - \frac{1}{B}\sum_{i=1}^{B}\cos(z_i,\; \bar{p}_{\neg m_k}),
\end{equation}
where $\bar{p}_{\neg m_k}$ is the global prototype of the missing modality ($\bar{p}_T$ for image-only, $\bar{p}_I$ for text-only). This pulls single-modality embeddings toward the complementary modality's global semantics.

\textbf{Regularization.} A prototype-proximal regularizer prevents local drift:
\begin{equation}\label{eq:reg}
\mathcal{L}_{\mathrm{reg}} = \frac{\lambda_a}{2}\!\sum_{v \in \mathcal{M}_k}\!\left(1 - \cos(p_v^k, \bar{p}_v)\right) + \lambda_s \left\| \theta_k - \theta \right\|_2^2,
\end{equation}
where $\mathcal{M}_k$ is the set of locally available modalities. The total client loss is:
\begin{equation}\label{eq:client_loss}
\mathcal{L}_k = \begin{cases}
\mathcal{L}_{\mathrm{NCE}} + \mathcal{L}_{\mathrm{reg}}, & \text{if paired}, \\
\lambda_{\mathrm{anc}} \cdot \mathcal{L}_{\mathrm{anchor}} + \mathcal{L}_{\mathrm{reg}}, & \text{otherwise}.
\end{cases}
\end{equation}

\textbf{Modality-Aware Statistics Reporting.}
After training, client $k$ reports a statistics vector $s_k$. For available modalities, the client computes a local prototype by averaging and normalizing embeddings. For the missing modality, the corresponding prototype slot is filled with the server's global EMA prototype $\bar{p}_{\neg m_k}$ received at the start of the round. A binary modality mask $m_k \in \{0,1\}^2$ indicates which prototypes are locally computed vs.\ filled. Concretely:
\begin{equation}\label{eq:stats}
s_k = \begin{cases}
(p_I^k, p_T^k, \gamma_k, m_k\!=\![1,1], \ell_k), & \text{if paired}, \\
(p_I^k, \bar{p}_T, \gamma_k, m_k\!=\![1,0], \ell_k), & \text{if image-only}, \\
(\bar{p}_I, p_T^k, \gamma_k, m_k\!=\![0,1], \ell_k), & \text{if text-only},
\end{cases}
\end{equation}
where $\gamma_k \in \R^4$ captures update geometry. This design ensures that all clients report a complete statistics vector while the mask allows downstream components to distinguish real from filled entries.

\subsection{Retrieval-Centric Semantic Router}
\label{sec:sec:router}

Standard FedAvg weights clients by dataset size, ignoring retrieval quality. RCSR introduces a learnable router that determines aggregation weights from retrieval-specific client statistics.

\textbf{Architecture.} The router $R_\phi$ encodes each $s_k$ into a token via a two-layer MLP, applies $L$ self-attention layers with modality-masked attention (padding slots for filled prototypes receive zero attention), and produces aggregation weights:
\begin{equation}\label{eq:router_weights}
w^R = \mathrm{softmax}\!\left(R_\phi(s_1, \ldots, s_K)\right) \in \Delta^{K-1}.
\end{equation}

\textbf{Prototype Consistency.} The router is trained to produce weights such that the weighted combination of client prototypes approximates the global EMA prototypes:
\begin{equation}\label{eq:proto_ent}
\mathcal{L}_{\mathrm{pc}} = \sum_{v \in \{I,T\}}\!\beta_v\!\left(1 - \cos\!\left(\textstyle\sum_k w_k^R p_v^k,\; \mathrm{sg}[\bar{p}_v]\right)\right) - \beta_4 H(w^R),
\end{equation}
where $\mathrm{sg}[\cdot]$ denotes stop-gradient on the EMA prototypes to prevent circular feedback, and $H(w^R)$ encourages diverse weight assignments.

\textbf{Alignment Consistency.} We define each client's alignment direction as $d_k = \mathrm{normalize}(p_I^k - p_T^k)$. For paired clients this captures the local image--text geometry; for image-only clients (where $p_T^k = \bar{p}_T$), $d_k$ reflects how far the local image distribution deviates from the global text anchor, which is itself a retrieval-relevant signal. The alignment consistency loss is:
\begin{equation}\label{eq:align_consist}
\mathcal{L}_{\mathrm{ac}} = 1 - \cos\!\left(\hat{d},\; \mathrm{sg}[\bar{d}]\right) + \sum_k w_k^R \left(1 - \cos(d_k, \hat{d})\right),
\end{equation}
where $\hat{d} = \mathrm{normalize}(\sum_k w_k^R d_k)$ and $\bar{d} = \mathrm{normalize}(\bar{p}_I - \bar{p}_T)$.

\textbf{Stabilization.} We apply two mechanisms to prevent feedback loops: \emph{(i)} stop-gradient on EMA prototypes $\bar{p}_v$ and global alignment $\bar{d}$ in all router losses (Eq.~\ref{eq:proto_ent}--\ref{eq:align_consist}), so the router cannot trivially match targets it also updates; \emph{(ii)} a warm-up phase of $T_w$ rounds using FedAvg before activating the router, ensuring that EMA prototypes are sufficiently reliable.

\textbf{Total Router Loss.} $\mathcal{L}_R = \mathcal{L}_{\mathrm{pc}} + \beta_5 \mathcal{L}_{\mathrm{ac}}$, trained on the server via backpropagation through the differentiable weight computation.

\subsection{Robust-Fair Aggregation}
\label{sec:fairness}

The router optimizes retrieval quality but does not guarantee fairness. We formulate a minimax game:
\begin{equation}\label{eq:minimax}
\min_\theta \max_{q \in \Delta^{N-1}} \sum_{k=1}^{N} q_k \, \tilde{\mathcal{L}}_k(\theta) + \tau_{\mathrm{fair}} H(q),
\end{equation}
where $\tilde{\mathcal{L}}_k = \ell_k / \bar{\ell}_{m_k}$ is the loss normalized by the running average loss of client $k$'s modality group (paired, image-only, or text-only). This normalization ensures that the fairness update is not dominated by clients with inherently larger loss scales. The adversary $q$ is updated via exponentiated gradient: $q_k \leftarrow q_k \cdot \exp(\eta_q \tilde{\ell}_k)$, then renormalized. The final aggregation weights fuse routing and fairness:
\begin{equation}\label{eq:fusion}
w_k = \frac{w_k^R \cdot q_k}{\sum_{j \in \mathcal{S}_t} w_j^R \cdot q_j}.
\end{equation}
The global model and prototypes are updated as $\theta_t = \theta_{t-1} + \sum_k w_k \Delta_k$ and $\bar{p}_v \leftarrow \mu \bar{p}_v + (1-\mu) \sum_k w_k p_v^k$.

\subsection{Client-Specific Adapter Personalization}
\label{sec:personalization}

While RCSR improves average retrieval quality across heterogeneous clients, a single shared model may not optimally serve all participants given their diverse data distributions and modality availability. We introduce per-client personalization adapters whose adaptation intensity is guided by the router's aggregation weights, yielding \textbf{RCSR-p}.

\textbf{Adapter Architecture.}
Each client $k$ maintains a lightweight personalization adapter $A_k\colon \R^d \to \R^d$ applied as a residual correction to the shared encoder output:
\begin{equation}\label{eq:personalized}
\tilde{z}_k = z + s_k \cdot A_k(z), \quad A_k(z) = W_2\,\mathrm{GELU}(W_1 z),
\end{equation}
where $W_2$ is initialized to near-zero so that $A_k$ begins as an approximate identity, ensuring minimal disruption to the global representation at personalization onset. The adapter is applied independently to image and text embeddings.

\textbf{Router-Guided Personalization Strength.}
The scalar $s_k \in (0, 0.5]$ ties adaptation intensity to the router weight $w_k^R$, which quantifies client $k$'s contribution to global retrieval coherence. Clients the router down-weights---due to alignment drift, limited paired data, or conflicting alignment directions---benefit more from local adaptation:
\begin{equation}\label{eq:strength}
s_k = \min\!\bigl(\lambda_p \cdot (1 + \max(0,\;0.5 - w_k^R)),\;0.5\bigr),
\end{equation}
where $\lambda_p$ is a base strength hyperparameter. This inverse coupling ensures that clients whose updates are already trusted globally receive only mild local correction, while clients underserved by the global model receive proportionally stronger compensation. Crucially, this signal is derived from existing router computation with no additional communication.

\textbf{Personalization Training.}
Personalization begins at round $T_p$ once the global model has stabilized. After each standard local training step, client $k$ performs additional adapter fine-tuning at a reduced learning rate $(0.1\eta)$. Paired clients minimize $\mathcal{L}_{\mathrm{NCE}}$ (Eq.~\ref{eq:infonce}) computed on personalized features $\tilde{z}_{k,I},\tilde{z}_{k,T}$; single-modality clients anchor their personalized features to the complementary global prototype via $\mathcal{L}_{\mathrm{anchor}}$ (Eq.~\ref{eq:anchor}). Adapter parameters $\{A_k\}$ are \emph{never} aggregated to the server---only the global shared adapters and projection heads participate in federated aggregation---keeping per-client personalization entirely local and communication-free.

\subsection{Theoretical Analysis}
\label{sec:theory}

\textbf{Geometry of cosine anchoring.}
For $\|a\|_2=\|b\|_2=1$, we have $1-\cos(a,b)=\frac{1}{2}\|a-b\|_2^2$.
Thus the anchoring loss pulls embeddings toward the opposite-modality global prototype on the unit sphere, keeping single-modality clients compatible with the shared cross-modal geometry.

\textbf{Robust envelope.}
Define the entropy-regularized robust objective:
\begin{equation}
\Phi(\theta)
=
\max_{q\in\Delta^{N-1}}
\left(
\sum_{k=1}^N q_k F_k(\theta)
+
\tau H(q)
\right),
\quad
H(q)=-\sum_{k=1}^N q_k\log q_k .
\label{eq:robust-objective}
\end{equation}

\textbf{Convergence guarantee.}
Under standard assumptions (smoothness, unbiased stochastic gradients, bounded heterogeneity, controlled adaptive aggregation), the outer minimization converges to a first-order stationary point at the standard $\mathcal{O}(1/T)$ rate:
\begin{equation}
\frac{1}{T}\sum_{t=0}^{T-1}
\mathbb{E}\!\left[\|\nabla F(\theta_t)\|_2^2\right]
\le
\frac{2(F(\theta_0)-F^\star)}{\gamma \eta E\, T}
+
C_4\bigl(\eta E + \sigma^2 + \eta E B^2\bigr)
+
C_5 \gamma \eta E,
\label{eq:convergence-bound}
\end{equation}
for constants $C_4, C_5 > 0$.
The complete proof, including all auxiliary lemmas and the personalization justification, is provided in Appendix~\ref{app:supp_theory}.

\textbf{Product-fusion interpretation.}
The fused weights $w_k \propto w_k^R q_k$ are the unique solution of $\arg\min_{p \in \Delta^{K-1}} \mathrm{KL}(p \| \pi)$, where $\pi_k = w_k^R q_k / \sum_j w_j^R q_j$. This gives a precise product-fusion view: $w^R$ down-weights conflicting updates, while $q$ up-weights hard clients, and the normalized product balances both.

\begin{table*}[t]
\centering
\caption{Main results under 50\% missing modality rate with 50\% client participation across four benchmarks. All methods trained with 30 clients for 200 rounds ($\alpha=0.1$).}
\label{tab:main_results_v3}
\resizebox{\textwidth}{!}{
\begin{tabular}{l|cc|cc|cc|cc|cc|cc|cc|cc}
\toprule
\multirow{3}{*}{\textbf{Method}} & \multicolumn{4}{c|}{\textbf{MS-COCO}} & \multicolumn{4}{c|}{\textbf{Flickr30K}} & \multicolumn{4}{c|}{\textbf{IAPR TC-12}} & \multicolumn{4}{c}{\textbf{MSR-VTT}} \\
\cmidrule(lr){2-5} \cmidrule(lr){6-9} \cmidrule(lr){10-13} \cmidrule(lr){14-17}
& \multicolumn{2}{c|}{I2T} & \multicolumn{2}{c|}{T2I} & \multicolumn{2}{c|}{I2T} & \multicolumn{2}{c|}{T2I} & \multicolumn{2}{c|}{I2T} & \multicolumn{2}{c|}{T2I} & \multicolumn{2}{c|}{I2T} & \multicolumn{2}{c}{T2I} \\
\cmidrule(lr){2-3} \cmidrule(lr){4-5} \cmidrule(lr){6-7} \cmidrule(lr){8-9} \cmidrule(lr){10-11} \cmidrule(lr){12-13} \cmidrule(lr){14-15} \cmidrule(lr){16-17}
& R@1 & R@5 & R@1 & R@5 & R@1 & R@5 & R@1 & R@5 & R@1 & R@5 & R@1 & R@5 & R@1 & R@5 & R@1 & R@5 \\
\midrule
FedAvg~\cite{mcmahan2017communication} & 35.26 & 63.68 & 36.14 & 63.84 & 36.46 & 67.51 & 38.57 & 68.07 & 28.43 & 52.36 & 29.78 & 53.42 & 24.12 & 45.28 & 25.48 & 46.82 \\
FedProx~\cite{li2020federated} & 36.18 & 64.52 & 37.22 & 64.96 & 37.92 & 68.74 & 39.84 & 69.23 & 29.26 & 53.18 & 30.64 & 54.36 & 24.86 & 46.12 & 26.24 & 47.68 \\
MFCPL~\cite{le2025cross} & 35.84 & 64.12 & 36.68 & 64.48 & 38.42 & 69.15 & 40.15 & 70.28 & 28.92 & 52.74 & 30.28 & 53.94 & 25.14 & 46.38 & 26.52 & 47.92 \\
CreamFL\cite{yu2023multimodal} & 37.42 & 65.86 & 38.58 & 66.18 & 40.23 & 70.58 & 41.65 & 71.02 & 30.48 & 54.62 & 31.86 & 55.84 & 26.18 & 47.42 & 27.64 & 48.96 \\
\textbf{RCSR} & \textbf{38.26} & \textbf{66.24} & \textbf{39.46} & \textbf{66.92} & \textbf{46.78} & \textbf{75.59} & \textbf{47.94} & \textbf{74.83} & \textbf{33.74} & \textbf{57.28} & \textbf{35.16} & \textbf{58.46} & \textbf{28.24} & \textbf{50.82} & \textbf{30.28} & \textbf{51.94} \\
\midrule
FedPer~\cite{arivazhagan2019federated} & 32.84 & 61.24 & 33.68 & 61.82 & 35.25 & 64.80 & 36.75 & 65.91 & 26.48 & 50.24 & 27.82 & 51.38 & 22.36 & 43.28 & 23.72 & 44.64 \\
MOON~\cite{li2021model} & 34.12 & 62.46 & 35.24 & 62.94 & 36.85 & 66.43 & 38.25 & 67.55 & 27.64 & 51.28 & 28.96 & 52.46 & 23.42 & 44.36 & 24.86 & 45.78 \\
FedMEKT~\cite{le2025fedmekt} & 34.86 & 63.12 & 35.92 & 63.58 & 37.46 & 67.85 & 39.15 & 68.97 & 28.18 & 51.86 & 29.53 & 53.12 & 23.94 & 44.99 & 25.38 & 46.34 \\
pFedMMA~\cite{ghiasvand2025pfedmma} & 36.95 & 64.85 & 38.17 & 66.25 & 40.22 & 72.45 & 42.07 & 71.28 & 30.12 & 54.28 & 31.48 & 55.52 & 26.04 & 47.28 & 27.48 & 48.82 \\
\textbf{RCSR-p} & \textbf{39.12} & \textbf{67.48} & \textbf{40.38} & \textbf{68.35} & \textbf{48.62} & \textbf{77.83} & \textbf{50.18} & \textbf{77.24} & \textbf{35.06} & \textbf{59.54} & \textbf{38.74} & \textbf{60.82} & \textbf{29.86} & \textbf{52.68} & \textbf{30.42} & \textbf{52.24} \\
\bottomrule
\end{tabular}
}
\end{table*}

\section{Experiment}
\label{sec:experiments}

We evaluate our method on four widely-used cross-modal retrieval benchmarks. Our experiments address three key questions: Does RCSR outperform federated baselines under various missing modality rates?  How does RCSR and its personalized component scale with the number of clients and severity of data heterogeneity? What is the contribution of each component?

\subsection{Experimental Setup}

\textbf{Datasets.}
We use four cross-modal retrieval benchmarks covering image--text and video--text tasks (Table~\ref{tab:datasets}).

\begin{table}[h]
\centering
\caption{Dataset statistics. Train/val/test splits follow standard protocols.}
\label{tab:datasets}
\begin{tabular}{lcccc}
\toprule
\textbf{Dataset} & \textbf{Train} & \textbf{Val} & \textbf{Test} & \textbf{Modality} \\
\midrule
Flickr30K~\cite{plummer2015flickr30k} & 29,783 & 1,000 & 1,000 & Image--Text \\
MS-COCO~\cite{lin2014microsoft} & 113,287 & 5,000 & 5,000 & Image--Text \\
IAPR TC-12~\cite{Grubinger2006IAPRTC12} & 16,000 & 2,000 & 2,000 & Image--Text \\
MSR-VTT~\cite{Xu_2016_CVPR} & 6,717 & 497 & 2,990 & Video--Text \\
\bottomrule
\end{tabular}
\end{table}

Flickr30K and MS-COCO use the Karpathy split~\cite{karpathy2015visualizing}, IAPR TC-12 uses the standard 80/20 split, and MSR-VTT uses the standard 7K split with 12 frames sampled per video.

\textbf{Data Partitioning.}
Since Flickr30K, MS-COCO, and IAPR TC-12 are caption datasets without native class labels, we derive pseudo-labels for Dirichlet partitioning via three steps: (1) encode all training captions with the frozen CLIP ViT-B/32 text encoder, yielding 512-dimensional embeddings; (2) cluster embeddings into $K\!=\!50$ semantic groups via K-Means ($\texttt{n\_init}\!=\!10$); (3) for each client, sample a distribution over $K$ clusters from $\mathrm{Dir}(\alpha)$ and assign samples proportionally. Lower $\alpha$ concentrates each client's data on fewer semantic topics. For MSR-VTT, the same procedure is applied to frame captions.

Under missing rate $\rho\!=\!50\%$, half of clients are designated single-modality (equal probability image-only or text-only). The assignment is \emph{client-level}: all samples on a single-modality client share the same modality mask. Partition statistics ($\alpha\!=\!0.1$, $N\!=\!30$): average samples per client are Flickr30K $\approx$ 967, MS-COCO $\approx$ 3,776, IAPR TC-12 $\approx$ 533, MSR-VTT $\approx$ 180. Pairwise JS-divergence between client distributions is $0.42 \pm 0.15$.

\textbf{Architecture and Training.}
We use frozen CLIP ViT-B/32~\cite{radford2021learning} (151M params) and only train bottleneck adapters (dim 64, 12 blocks, both encoders, 1.98M params) and projection heads (two-layer MLP $512 \to 512 \to 512$ per branch, 1.05M params), totaling 3.03M trainable parameters ($<$2\% of full model). We train 200 rounds with 50\% client participation, 1 local epoch, batch 128, lr $10^{-4}$ with cosine annealing and 5-round warmup, using AdamW with weight decay 0.01 on NVIDIA A800 GPUs.

\textbf{Baselines.}
Eight methods in three groups: \textit{(i)} general FL: FedAvg (weighted averaging by dataset size), FedProx (proximal regularizer $\mu\!=\!0.1$); \textit{(ii)} multimodal FL: MFCPL (cross-prototype construction, 10 prototypes per modality, CMPC+CMPR+CMA losses), CreamFL (representation-level ensemble distillation with contrastive regularization, EMA decay 0.9), FedMEKT (embedding knowledge transfer via KL divergence, temperature 3.0); \textit{(iii)} personalized FL: FedPer (adapters/projection heads as personalized layers, not aggregated), MOON (model-level contrastive learning, temperature 0.5, weight 1.0), pFedMMA (multi-modal adapter with shared+client-specific components, scale 0.5). All baselines share the same frozen CLIP + adapter + projection head architecture. Single-modality baselines use anchoring loss (Eq.~\ref{eq:anchor}). For each baseline, we adopt original hyperparameters and perform grid search when adaptation is required.

\textbf{Fairness Weight Normalization.}
The minimax fairness update uses normalized losses $\tilde{\mathcal{L}}_k = \ell_k / \bar{\ell}_{m_k}$, where $\bar{\ell}_{m_k}$ is the EMA ($\beta\!=\!0.9$) of the mean loss of client $k$'s modality group. This ensures the fairness update is not dominated by clients with inherently larger loss scales: paired clients have loss scale $\sim$1--5 while single-modality clients have scale $\sim$0.1--0.5. Without normalization, the exponentiated gradient would systematically upweight paired clients.

\textbf{Metrics.}
We report Recall@$K$ for $K \in \{1, 5, 10\}$ on both I2T and T2I retrieval. Mean Recall $\bar{R}@K = (R_{\text{I2T}}@K + R_{\text{T2I}}@K) / 2$ summarizes bidirectional performance. For fairness, we measure standard deviation of per-client R@1, worst-case client R@1, and performance gap.

\textbf{Hyperparameters.}
Table~\ref{tab:hyperparams} lists the complete settings.

\begin{table}[h]
\centering
\caption{Complete hyperparameter settings.}
\label{tab:hyperparams}
\resizebox{\columnwidth}{!}{%
\begin{tabular}{llc}
\toprule
\textbf{Category} & \textbf{Parameter} & \textbf{Value} \\
\midrule
\multirow{5}{*}{Training} & Communication rounds & 200 \\
& Client participation rate & 50\% \\
& Local epochs & 1 \\
& Batch size & 128 \\
& Optimizer / weight decay & AdamW / 0.01 \\
\midrule
\multirow{3}{*}{Learning Rate} & Initial LR & $10^{-4}$ \\
& Scheduler & Cosine annealing \\
& LR warm-up rounds & 5 \\
\midrule
\multirow{4}{*}{Architecture} & CLIP backbone & ViT-B/32 (frozen, 151M) \\
& Adapter bottleneck dim & 64 \\
& Adapter blocks & 12 (each transformer block, both encoders) \\
& Projection head & 2-layer MLP, $512 \to 512 \to 512$, ReLU \\
\midrule
\multirow{4}{*}{Client Loss} & $\lambda_{\mathrm{align}}$ (contrastive weight) & 0.1 \\
& $\lambda_s$ (proximal coefficient) & 0.01 \\
& $\lambda_{\mathrm{anc}}$ (anchoring weight) & 1.0 \\
& $\tau_{\mathrm{NCE}}$ (InfoNCE temperature) & 0.07 \\
\midrule
\multirow{6}{*}{Router} & Transformer layers / heads / dim & 2 / 4 / 128 \\
& $\beta_1, \beta_2$ (prototype consistency) & 1.0, 1.0 \\
& $\beta_4$ (entropy weight) & 0.2 \\
& $\beta_5$ (alignment consistency) & 0.3 \\
& Router LR & $10^{-3}$ \\
& Warm-up rounds $T_w$ & 20 \\
& EMA momentum $\mu$ & 0.9 \\
\midrule
\multirow{3}{*}{Fairness} & $\tau_{\mathrm{fair}}$ (entropy regularizer) & 0.1 \\
& $\eta_q$ (exponentiated gradient LR) & 0.1 \\
& Normalization momentum $\beta$ & 0.9 \\
\midrule
\multirow{2}{*}{Hardware} & GPU & NVIDIA A800 80GB \\
& Framework & PyTorch 2.0, CUDA 11.8 \\
\bottomrule
\end{tabular}%
}
\end{table}

\subsection{Main Results}

Table~\ref{tab:main_results_v3} presents the main comparison under the most challenging setting ($\alpha=0.1$, 50\% missing modality rate) across four cross-modal retrieval benchmarks, including both image-text and video-text tasks. The upper section compares general and multimodal FL methods, while the lower section compares personalized FL methods.

\textbf{Key Observations:} (1) RCSR consistently outperforms all baselines across all four datasets. The improvement over the strongest baseline CreamFL reaches 6.55\% R@1 on Flickr30K, demonstrating that retrieval-centric routing is particularly effective when data heterogeneity is severe. (2) Gains are well-balanced between I2T and T2I directions across all datasets, indicating that the router treats both modalities symmetrically. (3) RCSR-p further improves upon RCSR through personalization, achieving the largest gain of 8.77\% R@1 over pFedMMA on Flickr30K, confirming that routing and personalization are complementary. (4) The consistent improvements on MSR-VTT, a video-text dataset, show that RCSR generalizes beyond image-text retrieval to heterogeneous visual inputs.

\begin{figure}[h]
    \centering
    \includegraphics[width=1\linewidth, height=5.2cm]{fairness_heterogeneity.pdf}
    \caption{(a) Fairness ($\downarrow$, std. of per-client R@1) on Flickr30K with varying number of clients. (b)--(d) Personalized retrieval performance (R@1) on MSR-VTT, comparing RCSR-p against personalized FL baselines under varying client numbers, missing modality rates, and non-IID degrees. Dashed vertical lines indicate the default setting.}
    \label{fig:fairness_heter}
\end{figure}

\begin{figure*}[t]
\centering
\includegraphics[width=\textwidth, height=2.8cm]{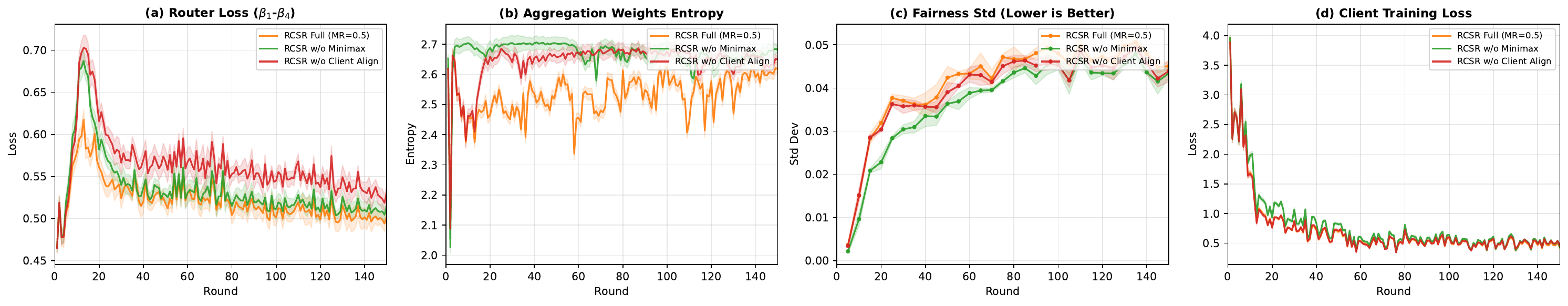}
\caption{Training dynamics of RCSR variants on Flickr30K dataset (30 clients, $\alpha=0.1$, MR=50\%). (a) Router loss converges stably. (b) Aggregation weights entropy shows non-uniform client weighting. (c) Fairness standard deviation demonstrates the impact of minimax game. (d) Client training loss decreases smoothly across all variants.}
\label{fig:training_dynamics}
\end{figure*}

\subsection{Robustness and Generalization Analysis}

\begin{figure*}[h]
\centering
\includegraphics[width=1\textwidth]{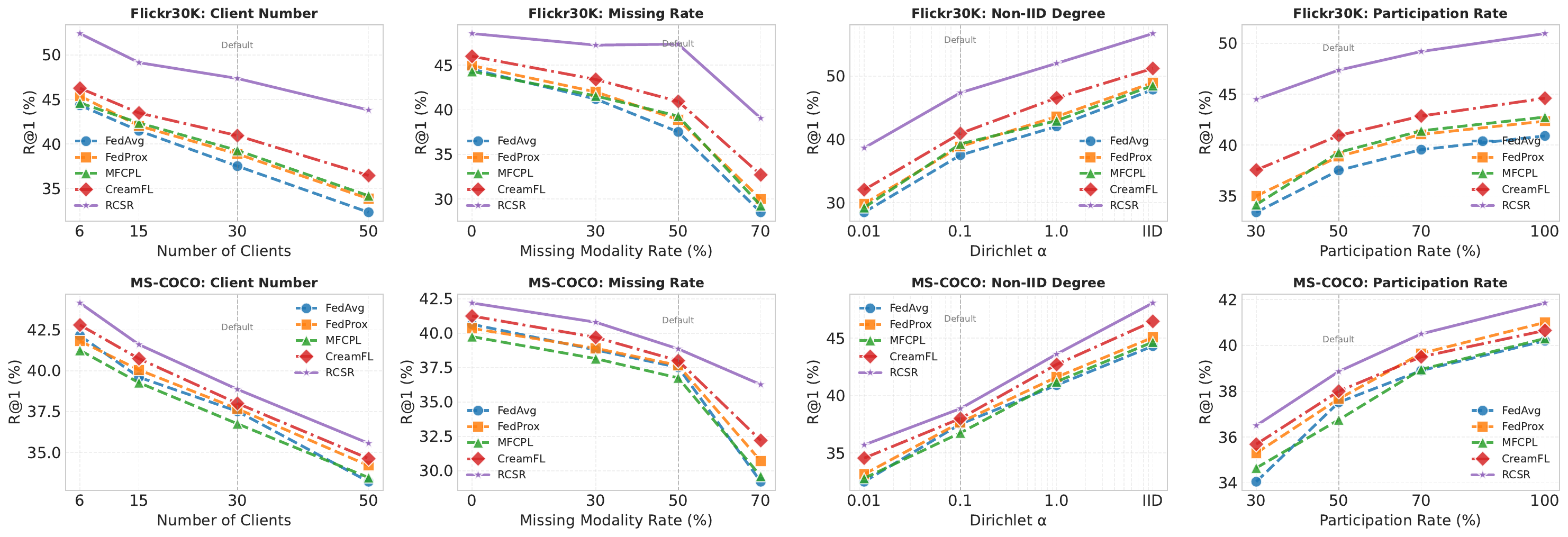}
\caption{Robustness analysis on Flickr30K (top) and MS-COCO (bottom). We vary four factors: number of clients, missing modality rate, non-IID degree, and client participation rate. Dashed vertical lines indicate the default setting. RCSR consistently outperforms all baselines, with the advantage amplifying under more challenging conditions.}
\label{fig:heter}
\end{figure*}

Figure~\ref{fig:heter} presents a comprehensive evaluation of RCSR across four dimensions---number of clients, missing modality rate, non-IID degree, and client participation rate---on both Flickr30K and MS-COCO. We compare against FedAvg, FedProx, MFCPL, and CreamFL under the default setting ($N=30$, $\alpha=0.1$, MR=50\%, 200 rounds) and vary one factor at a time.

\textbf{Client Scalability.} As the number of clients increases from 6 to 50, all methods degrade due to fewer samples per client, yet RCSR maintains the largest margin over the strongest baseline CreamFL, growing from 6.1\% to 7.4\% R@1 on Flickr30K. This indicates that retrieval-centric routing scales more gracefully than uniform aggregation, as the router adaptively concentrates communication budget on clients with the most informative updates.

\textbf{Missing Modality Rate.} RCSR exhibits superior robustness to modality incompleteness. On Flickr30K, RCSR drops only 1.2\% R@1 from MR=0\% to MR=50\%, whereas CreamFL degrades by 5.1\%. At the extreme MR=70\%, the gap between RCSR and CreamFL reaches 6.4\%. This confirms that the combination of anchoring loss and prototype-based routing effectively preserves cross-modal alignment even when the majority of clients lack paired data.

\textbf{Non-IID Degree.} We vary the Dirichlet concentration $\alpha$ from 0.01 (extreme heterogeneity) to 10.0 (near-IID). RCSR consistently outperforms all baselines across the entire spectrum. The advantage is most pronounced under severe heterogeneity ($\alpha=0.01$), where RCSR leads CreamFL by 6.6\% R@1 on Flickr30K. Even under near-IID conditions, RCSR maintains a 5.5\% gap, suggesting that the router provides benefits beyond handling data skew.

\textbf{Participation Rate.} RCSR remains effective even with low client participation. At only 30\% participation, RCSR achieves 44.5\% R@1 on Flickr30K---nearly matching CreamFL at 100\% participation (44.6\%). This practical advantage stems from the router's ability to identify and prioritize the most valuable clients in each round, making efficient use of limited participation budgets.

\subsection{Fairness and Personalization}

Beyond aggregate performance, we evaluate RCSR's fairness across clients and the personalization capability of RCSR-p on the video-text benchmark MSR-VTT. Figure~\ref{fig:fairness_heter} reports the results.

\textbf{Fairness.} Figure~\ref{fig:fairness_heter}(a) measures the standard deviation of per-client R@1 on Flickr30K as the number of clients scales from 6 to 50. RCSR achieves the lowest fairness std across all settings (0.0382 at 30 clients vs.\ CreamFL's 0.0442), confirming that the minimax fairness game effectively reduces performance disparity among clients. Notably, the fairness gap between RCSR and baselines widens with more clients, indicating that the routing mechanism increasingly benefits from larger client pools to identify and protect underperforming clients.

\textbf{Personalized Retrieval on MSR-VTT.} Figures~\ref{fig:fairness_heter}(b)--(d) evaluate RCSR-p against four personalized FL methods on MSR-VTT under varying client numbers, missing modality rates, and non-IID degrees. RCSR-p consistently outperforms all personalization baselines. Under the default setting ($N=30$, MR=50\%, $\alpha=0.1$), RCSR-p achieves 30.14\% R@1, surpassing the strongest baseline pFedMMA (26.76\%) by 3.38\%. The advantage is robust across all conditions: at extreme heterogeneity ($\alpha=0.01$), RCSR-p leads pFedMMA by 3.13\%; at severe missing modality (MR=70\%), the gap reaches 2.90\%. These results demonstrate that routing-based personalization combined with prototype-guided adaptation is more effective than layer-wise splitting or contrastive regularization alone in multimodal federated retrieval.

\subsection{Communication Efficiency and Training Dynamics}

We analyze the communication overhead and convergence behavior of RCSR to verify its practical feasibility in resource-constrained federated settings.

\begin{table}[h]
\centering
\caption{Communication Cost vs. Performance Gain (Relative to FedAvg Baseline). Performance measured by average improvement of $\bar{R}$@1 and $\bar{R}$@5 across all datasets.}
\label{tab:cost_vs_performance}
\resizebox{\columnwidth}{!}{%
\begin{tabular}{l|ccccc}
\toprule
 & \textbf{FedAvg} & \textbf{FedProx} & \textbf{MFCPL} & \textbf{CreamFL} & \textbf{RCSR} \\
\midrule
Communication Cost & 100\% & 100\% & 102\% & 103\% & 101\% \\
Performance Gain & - & +2.72\% & +2.41\% & +6.53\% & +16.12\% \\
\midrule
 & \textbf{FedPer} & \textbf{MOON} & \textbf{FedMEKT} & \textbf{pFedMMA} & \textbf{RCSR-p} \\
\midrule
Communication Cost & 75\% & 106\% & 109\% & 101\% & 105\% \\
Performance Gain& -5.12\% & -2.54\% & -0.69\% & +3.92\% & +17.83\% \\
\bottomrule
\end{tabular}%
}
\end{table}

\textbf{Communication Cost.} Since RCSR freezes the CLIP backbone (151M parameters) and only transmits bottleneck adapters (1.98M) and projection heads (1.05M) per round, the per-round communication volume is merely 3.03M parameters ($\sim$24\,MB per client), less than 2\% of the full model. Table~\ref{tab:cost_vs_performance} quantifies the cost-performance tradeoff relative to the FedAvg baseline. RCSR achieves the highest performance gain (+16.12\%) at a communication cost comparable to FedAvg (101\%), yielding the best cost-performance ratio among all standard FL methods. Among personalized methods, RCSR-p delivers +17.83\% gain at 105\% cost, significantly outperforming pFedMMA (+3.92\% at 101\%). In contrast, FedMEKT and MOON incur higher overhead (109\% and 106\%) yet even degrade performance.

\begin{figure}[h]
    \centering
    \includegraphics[width=0.47\textwidth]{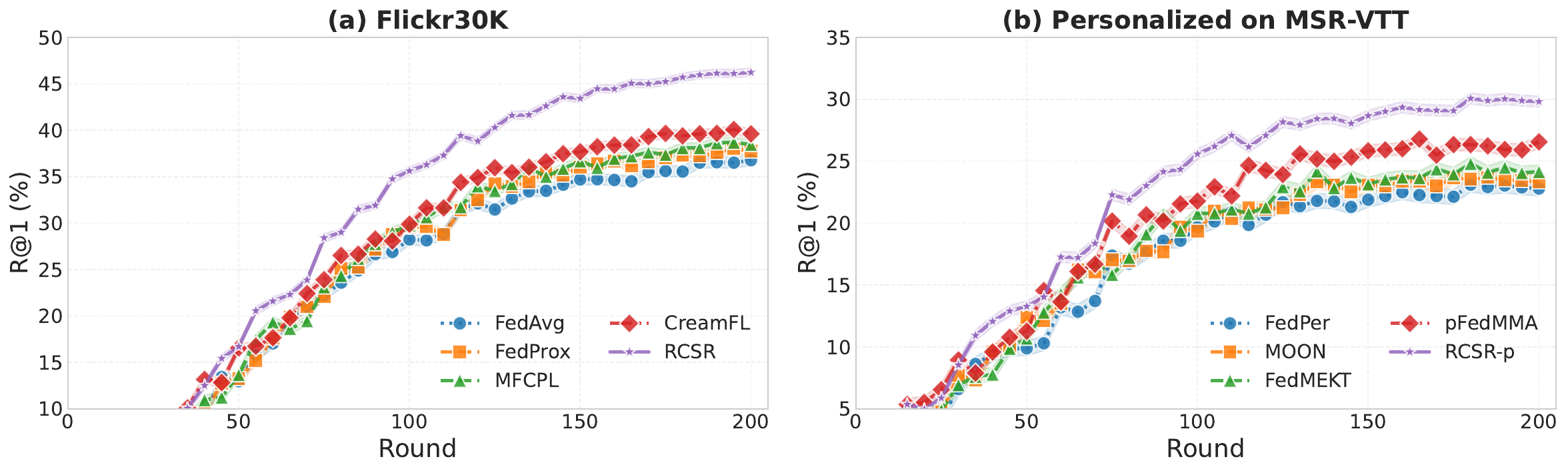}
    \caption{Training convergence curves on Flickr30K (a) and MSR-VTT (b). Shaded regions indicate confidence intervals. RCSR converges faster and to higher performance with lower variance than all baselines.}
    \label{fig:training_curves}
\end{figure}

\textbf{Convergence Behavior.} Figure~\ref{fig:training_curves} plots the training curves on Flickr30K (standard FL) and MSR-VTT (personalized FL). On Flickr30K, RCSR converges to 47.36\% R@1, substantially above CreamFL (40.94\%), with the gap emerging early and persisting throughout training. On MSR-VTT, RCSR-p reaches 30.14\% R@1, leading pFedMMA (26.76\%) by a similar margin. Both RCSR and RCSR-p exhibit smooth and stable convergence with lower variance (shaded confidence intervals), suggesting that the routing mechanism produces consistent aggregation across rounds without destabilizing oscillations.

\subsection{Ablation Study}

Table~\ref{tab:coco_ablation} presents the ablation study on MS-COCO under MR=50\% and $\alpha=0.1$, covering both component-level and personalization-level analysis. \textbf{Component Ablation.} Removing the client alignment loss causes the largest drop ($-$3.12\% $\bar{R}$@1), as the router depends on well-aligned local prototypes to produce meaningful aggregation weights. Removing the minimax fairness game also degrades performance ($-$2.08\% $\bar{R}$@1), since undertrained clients inject noise into the shared embedding space. Both components are complementary and neither can compensate for the other's absence. \textbf{Personalization Ablation.} We compare three variants to isolate the source of RCSR-p's improvement. Adding per-client adapters to FedAvg yields only modest gains (+0.89\% $\bar{R}$@1 over FedAvg), confirming that the adapter module alone is insufficient. Replacing router-weighted $\lambda$ with a uniform fixed strength degrades $\bar{R}$@1 by 0.44\%, demonstrating that coupling personalization intensity to the router's client-specific weights is essential. The full RCSR-p achieves the best result (39.75\% $\bar{R}$@1), validating that the routing signal uniquely enables effective personalization within this framework. Figure~\ref{fig:training_dynamics} further presents the ablation results of different components.

\begin{table}[t]
\centering
\caption{Ablation study on MS-COCO with 50\% missing modality rate. All variants trained with 30 clients for 150 rounds ($\alpha=0.1$). Best results in \textbf{bold}.}
\label{tab:coco_ablation}
\resizebox{\columnwidth}{!}{%
\begin{tabular}{l|ccc|ccc|c}
\toprule
\multirow{2}{*}{\textbf{Method Variant}} & \multicolumn{3}{c|}{\textbf{Image-to-Text}} & \multicolumn{3}{c|}{\textbf{Text-to-Image}} & \multirow{2}{*}{\textbf{$\bar{R}$@1}} \\
\cmidrule(lr){2-4} \cmidrule(lr){5-7}
& R@1 & R@5 & R@10 & R@1 & R@5 & R@10 & \\
\midrule

RCSR Full (Ours)          & \textbf{38.26} & \textbf{66.24} & \textbf{77.88} & \textbf{39.46} & \textbf{66.92} & \textbf{77.40} & \textbf{38.86} \\
\quad w/o Minimax Game    & 36.32          & 64.78          & 76.50          & 37.24          & 65.42          & 76.50          & 36.78 \\
\quad w/o Client Align    & 34.34          & 63.40          & 74.86          & 37.14          & 64.16          & 75.40          & 35.74 \\
\midrule

FedAvg + per-client adapter     & 36.48 & 65.24 & 75.68 & 37.28 & 65.46 & 75.85 & 36.88 \\
RCSR (no personalization)       & 38.26 & 66.24 & 77.88 & 39.46 & 66.92 & 77.40 & 38.86 \\
RCSR-p (fixed $\lambda$)        & 38.70 & 66.86 & 78.42 & 39.92 & 67.64 & 78.15 & 39.31 \\
\textbf{RCSR-p (router-weighted $\lambda$)} & \textbf{39.12} & \textbf{67.48} & \textbf{78.95} & \textbf{40.38} & \textbf{68.35} & \textbf{78.85} & \textbf{39.75} \\
\bottomrule
\end{tabular}%
}
\end{table}

\section{Discussion}

\textbf{Fairness and average performance are aligned.} The minimax game raises the worst-case client's absolute performance (0.45 vs.\ 0.43 for FedAvg at MR=50\%) rather than merely redistributing scores---an outcome we attribute to the geometric structure of retrieval tasks, where undertrained clients inject alignment noise that degrades the shared embedding space for all clients. We apply z-score normalization to fairness weights within each modality group to ensure comparability.

\textbf{Single-modality statistics construction.} Image-only clients compute $p_I^k$ from available images and use the broadcast global prototype $\bar{p}_T$ as $p_T^k$ (and vice versa for text-only clients). The modality mask $m_k$ indicates missing prototypes to the router, which masks the corresponding entries in $\mathcal{L}_{\mathrm{pc}}$. We compare three strategies---zero-padding ($\bar{R}$@1 = 36.07), learnable placeholder (37.70), and our global EMA fallback (38.86)---on MS-COCO and find that the EMA fallback provides the most informative cross-modal signal.

\textbf{Router stability.} The router depends on global EMA prototypes that are themselves updated using router weights. We stabilize training via warm-up (20 rounds of FedAvg before router activation), stop-gradient on $\bar{p}_v$ when computing $\nabla_\phi \mathcal{L}_{\mathrm{pc}}$, and high EMA momentum ($\mu=0.9$).

\textbf{When does personalization help most?} RCSR-p's gains over RCSR are largest for clients with low router weight (typically single-modality or highly non-IID clients). For these clients, the router already signals that their updates conflict with the global objective; stronger local adaptation (Eq.~\ref{eq:strength}) compensates by specializing the encoder output to their local distribution without affecting the globally aggregated parameters.

\textbf{Privacy and overhead.} Client prototypes and losses may leak distributional information; threat models and differentially private variants are discussed in Appendix~\ref{sec:privacy}. The router's forward pass processes $K=10$ client statistics ($<50$ms), negligible vs.\ local training ($\sim$2 min). Communication increases by only near 1\% ($\sim$100KB per client statistics vs.\ $\sim$12MB parameters). Per-client adapters add no communication cost.

\section{Conclusion}
We presented RCSR, a federated cross-modal retrieval framework that jointly addresses missing modalities, data heterogeneity, and client fairness. Prototype anchoring enables single-modality clients to contribute meaningful updates; the retrieval-centric semantic router adaptively weights aggregation based on cross-modal alignment quality; and minimax-game reweighting improves worst-case client performance while benefiting global accuracy. RCSR-p extends the framework with router-guided per-client adapters that provide communication-free local personalization. Across Flickr30K, MS-COCO, IAPR TC-12, and MSR-VTT, RCSR and RCSR-p consistently outperform both general and personalized FL baselines under severe non-IID and high missing-modality conditions, establishing a practical and communication-efficient foundation for privacy-preserving multimodal retrieval.

\bibliographystyle{ACM-Reference-Format}
\bibliography{sample-base}

\appendix
\settopmatter{printacmref=false}

\section{Convergence Analysis}
\label{app:supp_theory}

This section provides additional theoretical justification for the proposed method.
Our goal is not to introduce a separate optimization framework beyond the main paper,
but rather to show that the proposed adaptive aggregation and personalization mechanism
is compatible with a standard federated optimization view
and does not invalidate the usual convergence argument under mild assumptions.
Our analysis is conceptually connected to standard federated optimization
and personalized federated learning formulations
such as FedAvg~\cite{mcmahan2017communication},
FedProx~\cite{li2020fedprox},
SCAFFOLD~\cite{karimireddy2020scaffold},
Per-FedAvg~\cite{fallah2020personalized},
pFedMe~\cite{dinh2020personalized},
Ditto~\cite{li2021ditto},
FedPer~\cite{arivazhagan2019federated},
and FedRep~\cite{collins2021exploiting}.

\subsection{Problem Setup and Notation Mapping}

Consider the global objective
\begin{equation}
F(\theta) := \sum_{k=1}^{N} p_k F_k(\theta),
\qquad
\sum_{k=1}^{N} p_k = 1,\quad p_k \ge 0,
\end{equation}
where \(N\) is the total number of clients,
\(F_k\) corresponds to the local loss \(\mathcal{L}_k\) defined in Eq.~\eqref{eq:client_loss},
and \(p_k\) denotes the data-proportional reference weight used in standard FedAvg.

At communication round \(t\), the server broadcasts the global model \(\theta_t\).
Each participating client \(k\) initializes from \(\theta_t\) and performs \(E\) local update steps:
\begin{equation}
\theta_{k,t}^{e+1}
=
\theta_{k,t}^{e}
-
\eta \Bigl(
\nabla F_k(\theta_{k,t}^{e};\xi_{k,t}^{e})
+
\lambda_s(\theta_{k,t}^{e}-\theta_t)
\Bigr),
\qquad e=0,\dots,E-1.
\label{eq:supp_local_update}
\end{equation}

Define the local model displacement \(\Delta_k := \theta_{k,t}^{E} - \theta_t\).
The server update can be written as
\(\theta_{t+1} = \theta_t + \gamma \sum_{k=1}^{N}\alpha_{k,t}\Delta_k\),
where \(\gamma = 1\) and \(\alpha_{k,t}\) are the adaptive weights from Eq.~\eqref{eq:fusion}.

To make the role of adaptive weighting explicit, we rewrite this as
\begin{equation}
\theta_{t+1}
=
\theta_t + \gamma \sum_{k=1}^{N} p_k \Delta_k + \gamma e_t,
\label{eq:supp_server_update_error}
\end{equation}
where \(e_t := \sum_{k=1}^{N}(\alpha_{k,t}-p_k)\Delta_k\) is the aggregation perturbation introduced by the adaptive weighting rule.

\subsection{Assumptions}

We impose the following standard assumptions.

\begin{assumption}[Smoothness]
Each local objective \(F_k\) is \(L\)-smooth:
\(\|\nabla F_k(x)-\nabla F_k(y)\| \le L\|x-y\|, \quad \forall x,y \in \mathbb{R}^d\).
\end{assumption}

\begin{assumption}[Unbiased stochastic gradients]
\(\mathbb{E}_{\xi}[\nabla F_k(\theta;\xi)] = \nabla F_k(\theta)\) and \(\mathbb{E}_{\xi}[\|\nabla F_k(\theta;\xi)-\nabla F_k(\theta)\|^2] \le \sigma^2\).
\end{assumption}

\begin{assumption}[Bounded second moment]
\(\mathbb{E}[\|\nabla F_k(\theta;\xi)\|^2] \le G^2, \quad \forall k,\theta\).
\end{assumption}

\begin{assumption}[Bounded heterogeneity]
\(\sum_{k=1}^{N} p_k \|\nabla F_k(\theta)-\nabla F(\theta)\|^2 \le B^2, \quad \forall \theta\).
\end{assumption}

\begin{assumption}[Controlled adaptive aggregation]
There exists \(c_\alpha \ge 0\) such that \(\mathbb{E}[\|e_t\|^2] \le c_\alpha \sum_{k=1}^{N} \mathbb{E}[\|\Delta_k\|^2]\).
\end{assumption}

Assumptions 1--4 are standard in non-convex FL analysis. Assumption 5 holds for RCSR because: (i) softmax weights are bounded in $[0,1]$ and sum to 1, so by Cauchy--Schwarz, $\|e_t\|^2 \le N \sum_k \|\Delta_k\|^2$; (ii) the loss-normalized fairness update prevents unbounded $q_k$; (iii) warm-up and stop-gradient provide additional stability.

\subsection{Convergence Guarantee}

\begin{theorem}
Suppose Assumptions 1--5 hold and \(F^\star := \inf_{\theta}F(\theta)>-\infty\).
Then for sufficiently small \(\eta\) and \(\gamma\):
\begin{equation}
\frac{1}{T}\sum_{t=0}^{T-1}
\mathbb{E}\bigl[\|\nabla F(\theta_t)\|^2\bigr]
\le
\frac{2(F(\theta_0)-F^\star)}{\gamma \eta E\, T}
+
C_4\bigl(\eta E + \sigma^2 + \eta E B^2\bigr)
+
C_5 \gamma \eta E,
\end{equation}
for constants \(C_4,C_5>0\).
If \(\eta\) and \(\gamma\) are chosen so that residual terms vanish with \(T\), the iterates converge to a first-order stationary point.
\end{theorem}

\begin{proof}[Proof sketch]
By $L$-smoothness, \(F(\theta_{t+1}) \le F(\theta_t) + \langle \nabla F(\theta_t), \theta_{t+1}-\theta_t \rangle + \frac{L}{2}\|\theta_{t+1}-\theta_t\|^2\). Substituting Eq.~\eqref{eq:supp_server_update_error} and applying Young's inequality to both the residual $r_t$ (from the averaged local displacement) and the perturbation $e_t$ (from adaptive weighting), all higher-order terms are bounded by $\gamma \eta E (\eta E + \sigma^2 + \eta E B^2)$ and $\gamma^2 \eta^2 E^2$. Summing over $t=0,\ldots,T-1$ and rearranging yields the result.
\end{proof}

\subsection{Personalization Justification}

\begin{proposition}
Suppose \(F_k\) is \(L\)-smooth and client \(k\) performs \(E\) local adaptation steps from \(\theta_t\) with \(\eta \le 1/L\). Then:
\(F_k(\theta_{k,t}^{E}) \le F_k(\theta_t) - \frac{\eta}{2}\sum_{e=0}^{E-1}\|\nabla F_k(\theta_{k,t}^{e})\|^2\).
Hence the personalized model is no worse than the shared global initialization on the client's own objective.
\end{proposition}

Furthermore, the proximal formulation $\widetilde{F}_{k,t}(\theta) = F_k(\theta) + \frac{\lambda_s}{2}\|\theta-\theta_t\|^2$ ensures that $\|\theta_{k,t}^{\star}-\theta_t\|^2 \le \frac{2}{\lambda_s}(F_k(\theta_t)-F_k(\theta_{k,t}^{\star}))$, explicitly controlling deviation from the global model.

\begin{table*}[t]
\centering
\caption{Full results on Flickr30K and MS-COCO with R@1, R@5, R@10 ($N\!=\!30$, $\alpha\!=\!0.1$, MR=50\%). $\bar{R}$@1 in \textbf{bold}. Best in \underline{underline}.}
\label{tab:main_full}
\resizebox{0.8\textwidth}{!}{%
\small
\begin{tabular}{l|ccc|ccc|c|ccc|ccc|c}
\toprule
\multirow{3}{*}{\textbf{Method}} & \multicolumn{7}{c|}{\textbf{Flickr30K}} & \multicolumn{7}{c}{\textbf{MS-COCO}} \\
\cmidrule(lr){2-8} \cmidrule(lr){9-15}
& \multicolumn{3}{c|}{I2T} & \multicolumn{3}{c|}{T2I} & \multirow{2}{*}{$\bar{R}$@1}
& \multicolumn{3}{c|}{I2T} & \multicolumn{3}{c|}{T2I} & \multirow{2}{*}{$\bar{R}$@1} \\
\cmidrule(lr){2-4} \cmidrule(lr){5-7} \cmidrule(lr){9-11} \cmidrule(lr){12-14}
& R@1 & R@5 & R@10 & R@1 & R@5 & R@10 & & R@1 & R@5 & R@10 & R@1 & R@5 & R@10 & \\
\midrule
\multicolumn{15}{l}{\textit{Standard FL}} \\
FedAvg      & 36.46 & 67.51 & 79.25 & 38.57 & 68.07 & 79.42 & 37.52 & 35.26 & 63.68 & 75.42 & 36.14 & 63.84 & 74.96 & 35.70 \\
FedProx     & 37.92 & 68.74 & 80.15 & 39.84 & 69.23 & 80.68 & 38.88 & 36.18 & 64.52 & 76.18 & 37.22 & 64.96 & 76.45 & 36.70 \\
MFCPL       & 38.42 & 69.15 & 80.52 & 40.15 & 70.28 & 81.25 & 39.29 & 35.84 & 64.12 & 75.86 & 36.68 & 64.48 & 76.12 & 36.26 \\
CreamFL     & 40.23 & 70.58 & 81.85 & 41.65 & 71.02 & 82.15 & 40.94 & 37.42 & 65.86 & 77.24 & 38.58 & 66.18 & 77.52 & 38.00 \\
\textbf{RCSR} & \underline{46.78} & \underline{75.59} & \underline{85.42} & \underline{47.94} & \underline{74.83} & \underline{85.18} & \textbf{47.36} & \underline{38.26} & \underline{66.24} & \underline{77.88} & \underline{39.46} & \underline{66.92} & \underline{77.40} & \textbf{38.86} \\
\midrule
\multicolumn{15}{l}{\textit{Personalized FL}} \\
FedPer      & 35.25 & 64.80 & 76.92 & 36.75 & 65.91 & 77.48 & 36.00 & 32.84 & 61.24 & 73.56 & 33.68 & 61.82 & 73.28 & 33.26 \\
MOON        & 36.85 & 66.43 & 78.15 & 38.25 & 67.55 & 78.62 & 37.55 & 34.12 & 62.46 & 74.28 & 35.24 & 62.94 & 74.52 & 34.68 \\
FedMEKT     & 37.46 & 67.85 & 79.25 & 39.15 & 68.97 & 79.48 & 38.30 & 34.86 & 63.12 & 74.85 & 35.92 & 63.58 & 75.12 & 35.39 \\
pFedMMA     & 40.22 & 72.45 & 82.43 & 42.07 & 71.28 & 81.98 & 41.15 & 36.95 & 64.85 & 76.62 & 38.17 & 66.25 & 77.18 & 37.56 \\
\textbf{RCSR-p} & \underline{48.62} & \underline{77.83} & \underline{87.27} & \underline{50.18} & \underline{77.24} & \underline{86.85} & \textbf{49.40} & \underline{39.12} & \underline{67.48} & \underline{78.62} & \underline{40.38} & \underline{68.35} & \underline{78.85} & \textbf{39.75} \\
\bottomrule
\end{tabular}%
}
\end{table*}

\section{Privacy Considerations}
\label{sec:privacy}

Although federated learning avoids direct sharing of raw data, RCSR's transmitted signals (adapter updates, modality indicators, local prototypes, routing statistics, and training losses) may still support information leakage. We consider an honest-but-curious server threat model.

\textbf{Potential leakage sources.}
Local prototypes may reveal coarse semantic properties of a client's data distribution. Modality indicators may correlate with institutional capability or device type. Routing statistics (update geometry, alignment direction, loss) may reveal how difficult a client's task is or whether it belongs to a minority subgroup. Client-specific adapters, if exposed, could fingerprint local data.

\textbf{Mitigation strategies.}
RCSR is compatible with: (i) secure aggregation~\cite{bonawitz2017practical} so the server only sees aggregated results; (ii) gradient/prototype clipping with noise injection for differential privacy~\cite{geyer2017differential}; (iii) coarsened or encrypted modality indicators; (iv) minimum-support prototype sharing (upload only when local samples exceed a threshold); (v) keeping per-client adapters strictly local.

\textbf{Limitations.}
We do not claim differential privacy or provable resistance to reconstruction/inference attacks. A rigorous privacy analysis of prototype-routed multimodal federated retrieval is an important direction for future work. RCSR remains substantially more privacy-friendly than centralized training: raw data never leave the client, the CLIP backbone is frozen, and only lightweight parameters ($<$2\% of the model) are transmitted.

\section{Supplementary Experiments}
\label{app:experiments}

\subsection{Full Results with R@10}

Table~\ref{tab:main_full} reports R@1, R@5, and R@10 for all methods on Flickr30K and MS-COCO, providing the complete evaluation beyond the R@1 and R@5 reported in the main paper.

\subsection{Communication Cost Breakdown}

In each round, client $k$ uploads: (i) model update $\Delta_k$: 3.03M parameters $\times$ 4 bytes = 12.12\,MB; (ii) router statistics $s_k$: $\sim$10\,KB. The server downloads the global model (12.12\,MB) to each selected client. Total per client per round: $\sim$24.2\,MB. Full fine-tuning would transmit 154M parameters (616\,MB), so RCSR achieves a $50\times$ reduction.

\subsection{Computational Cost}

Local training takes $\sim$2 min per client per round. The router's forward and backward passes process $K=15$ client statistics in $<$150\,ms total. Server-side aggregation takes $<$10\,ms. Total server overhead is $<$1\% of local training. Total runtime: $\sim$45 min per run (30 clients, 200 rounds, 1 GPU).

\subsection{Single-Modality Statistics Construction}

We compare three strategies for filling the missing-modality prototype slot on MS-COCO ($N\!=\!30$, $\alpha\!=\!0.1$, MR=50\%): zero-padding ($\bar{R}$@1 = 36.07), learnable placeholder (37.70), and global EMA fallback (38.86). Zero-padding creates degenerate alignment directions; learnable placeholders are shared across clients and do not track the evolving global distribution; the EMA fallback uses the server's running average prototype, providing the most informative cross-modal signal.

\subsection{Reproducibility Checklist}

\begin{itemize}
    \item \textbf{Data partitioning}: CLIP text embeddings $\to$ K-Means ($K\!=\!50$, $\texttt{n\_init}\!=\!10$) $\to$ Dirichlet ($\alpha\!=\!0.1$), seed 42.
    \item \textbf{Missing modality}: Client-level, $\rho\!=\!50\%$, equal probability image-only/text-only, seed 42.
    \item \textbf{Training}: 200 rounds, 50\% participation, 1 local epoch, batch 128, lr $10^{-4}$, cosine annealing, 5-round warmup, AdamW, weight decay 0.01.
    \item \textbf{Architecture}: CLIP ViT-B/32 frozen, adapters (dim 64, each transformer block, both encoders), projection heads ($512 \to 512 \to 512$, ReLU). Initialization: adapter weights $\sim \mathcal{N}(0, 0.01)$, projection heads $\sim \mathcal{N}(0, 0.02)$.
    \item \textbf{Router}: 2 transformer layers, 4 heads, dim 128, lr $10^{-3}$, activated after $T_w\!=\!20$ rounds.
    \item \textbf{EMA}: $\mu\!=\!0.9$ for global prototypes.
    \item \textbf{Hardware}: NVIDIA A800 80GB, PyTorch 2.0, CUDA 11.8.
    \item \textbf{Runtime}: $\sim$45 min per run (30 clients, 200 rounds, 1 GPU).
\end{itemize}

\end{document}